\documentclass[runningheads]{llncs}
\usepackage{times}
\usepackage{graphicx}
\usepackage{latexsym}
\usepackage{titletoc}
\usepackage{times}  
\usepackage{helvet} 
\usepackage{courier}  
\usepackage[hyphens]{url}  
\usepackage[misc]{ifsym}
\usepackage{graphicx}  
\usepackage{cite}
\usepackage{amsmath,amssymb,amsfonts}
\usepackage{microtype}
\usepackage{algorithmic}
\usepackage{subfig}
\usepackage{textcomp}
\usepackage{fancyvrb}
\usepackage{xcolor}
\usepackage{multirow}
\usepackage{subfig}
\usepackage[normalem]{ulem}
\useunder{\uline}{\ul}{}

\usepackage{graphicx}
\begin{document}
\title{Multi-Source Anomaly Detection in Distributed IT Systems}
%
%
\author{Jasmin~Bogatinovski\inst{1,3}\and
Sasho~Nedelkoski\inst{1,3}\\
}
\authorrunning{Bogatinovski~J.~Nedelkoski~S.~et~al.}


%
\institute{Distributed Operating Systems, 
TU Berlin, Berlin, Germany \\
\email{\{jasmin.bogatinovski, nedelkoski\}@tu-berlin.de}\\
Equal contribution
}
\maketitle
\setcounter{footnote}{0}
\tocauthor{Sasho Nedelkoski \and Jasmin Bogatinovski}
\authorrunning{Bogatinovski J. Nedelkoski S.}
%
\institute{Princeton University, Princeton NJ 08544, USA \and
Springer Heidelberg, Tiergartenstr. 17, 69121 Heidelberg, Germany
\email{lncs@springer.com}\\
\url{http://www.springer.com/gp/computer-science/lncs} \and
ABC Institute, Rupert-Karls-University Heidelberg, Heidelberg, Germany\\
\email{\{abc,lncs\}@uni-heidelberg.de}}

\begin{abstract}
The multi-source data generated by distributed systems, provide a holistic description of the system. Harnessing the joint distribution of the different modalities by a learning model can be beneficial for critical applications for maintenance of the distributed systems. One such important task is the task of anomaly detection where we are interested in detecting the deviation of the current behaviour of the system from the theoretically expected.  In this work, we utilize the joint representation from the distributed traces and system log data for the task of anomaly detection in distributed systems. We demonstrate that the joint utilization of traces and logs produced better results compared to the single modality anomaly detection methods. Furthermore, we formalize a learning task - next template prediction NTP, that is used as a generalization for anomaly detection for both logs and distributed trace. Finally, we demonstrate that this formalization allows for the learning of template embedding for both the traces and logs. The joint embeddings can be reused in other applications as good initialization for spans and logs.
\end{abstract}

\section{Introduction}
The complexity of the multi-layered IT infrastructures such as the Internet of Things, distributed processing frameworks, databases and operating systems, is constantly increasing~\cite{NedelkoskiCLOUD}. To meet the consumers' expectations of fluent service with low response times guarantees and availability, the service providers highly rely on the high volumes of monitoring data. The massive volumes of data lead to maintenance overhead for the operators and require introducing of data-driven tools to process the data. 

A crucial task for such tools is to correctly identify the symptoms of deviation of the current behaviour system from the expected one. Due to the large volumes of data, the anomaly detector should produce a small number of false-positive alarms, thus reducing the efforts of the operators, while at the same time producing a high detection rate. The benefit of timely detection allows prevention of potential failures and increases the opportunity window for conducting a successful reaction from the operator. This is especially important if urgent expertise and/or administration activity is required. The symptoms often are notified whenever there are performance problems or system failures and usually manifests as some fingerprints within the monitored data: logs, metrics or distributed traces. 

The monitored system data represent the state of the system at any time point. They are grouped into three categories-modalities:  metrics,  application logs, and distributed traces~\cite{sridharan2018distributed}. The metrics are time-series data that represent the utilization of the available resources and the status of the infrastructure. Typically they involve measuring of the CPU, memory and disk utilization, as well as data as network throughput, and service call latency. Application logs are print statements appearing in code with semi-structured content. They represent interactions between data, files, services, or applications containing a rich representative structure on a service level. Service, microservices, and other systems generate logs which are composed of timestamped records. Distributed traces chains the service invocations as workflows of execution of HTTP or RPC requests. Each part of the chain in the trace is called an event or span. A property of this type of data is that it preserves the information for the execution graph on a (micro)service level. Thus, the information for the interplay between the components is preserved. 

The log data can produce a richer description on a service level since they are fingerprints of the program execution within the service. On the other side, the traces do not have much information on system-level information but preserve the overall graph of request execution. Referring to the different aspects of the system, the logs and traces provide orthogonal information for the distributed systems behaviour. Building on this observation in this work, we introduce an anomaly detection multi-source approach that can consider the data from both the traces and logs, jointly.  We demonstrate the usability of time-aligned log and tracing data to produce better results on the task of anomaly detection as compared to the single modalities as the main contribution to this work. The results show that the model build under the joint loss from both the logs and trace data can exploit some relationship between the modalities. The approach is trainable end-to-end and does not require the building of separate models for each of the modalities. As a second contribution, we consider the introduction of vector embeddings for the spans within the trace. The adopted approach allows the definition of the span vectors as a pooling over the words they are composed of. We refer to these vector embeddings as span2vec.

\section{Related Work}
The literature recognizes various approaches concerned with anomaly detection in distributed systems from single modalities. We review the single modalities approaches for both logs and traces. We also provide an overview of the existing multi-modal approaches, however, none of them jointly considers both traces and logs. 

The most common approaches for anomaly detection from log data roughly follows a two-step composition - log parsing followed by a method for anomaly detection. The first step allows for an appropriate log representation. One challenge during this procedure is the reduction of the noise in the log data. This noise in a log message is present due to the various parameters parts of the log can take during execution. To this end, there are many proposed techniques for log parsing \cite{Drain, LogSys, NuLog}. A detailed overview and comparison across benchmarks of these techniques are given in \cite{LogPAI}. After the template extraction, there are two general approaches to represent the logs. The first one is based on word frequencies and metrics derived from the logs (e.g TF-IDF) \cite{invarientMainng, PCA, ALogParsing, ICDM} or reusing word representation of the logs, based on corpora of words. The second approach aims at translating the templates into sequences of templates - most often represented as sequences of integers or sequences of vectors. Such representation allows modelling the sequential execution of a program workflow. One of the most commonly utilized approaches is RNN-based(e.g LSTM, GRU) \cite{DeepLog}. They often are coupled with an additional mechanism such as attention to allow for better preservation of the semantic information inside the logs \cite{LogAnomaly}. Depending on the data representation, various methods are utilized from both the supervised and unsupervised domains of machine learning. However, due to easier practical adoption and the absence of labels, the unsupervised methods are preferred. 

The available approaches for anomaly detection from tracing data are scarce. They usually model the normal execution of a workload, represented within the trace by utilizing history $h$ of recent trace events as input. They decompose the trace in its building blocks, the events/spans, and predict the next span in the sequence. The anomaly detection is done with imposing thresholds on the number of errors the LSTM is making for the corresponding trace predicted \cite{NedelkoskiCCGRID, NedelkoskiCLOUD}.
Further approaches aim to capture the execution of a complete workload into a finite state automata (FSA) \cite{FSA}. However, the FSA approaches are dependent on specific tracing implementation systems. The unification of this approach with other types of modalities such as the log data due to the assumed homogeneous structure of the states building the FSA is harder.

Several works on multi-modal learning for anomaly detection demonstrate the feasibility of using different modalities of data for anomaly detection \cite{Park, Srivastava}. In the context of large scale ICT systems, the authors in \cite{NedelkoskiCLOUD} consider the joint exploitation of traces and the corresponding response times of the spans within the trace. More specifically, a multi-modal LSTM-based method, trained jointly on both modalities is introduced, showing the additional value added by the shared information, improves the anomaly detection scores. In \cite{Japan} a Multimodal Variational Autoencoder approach is adopted for effectively learning the relationships among cross-domain data which provide good results for anomaly detection build on the logs and metrics as modalitites. However, they do not preserve the information for the overall microservice architecture.

To the best of our knowledge, the literature does not yet recognize methods for joint consideration of logs and traces as fundamentally complementary data sources describing the distributed IT systems. Hence in this work, we propose an approach on how to jointly consider the complement information within the logs and traces.

\section{Multimodal approach for anomaly detection from heterogeneous data}
In this section, we describe the multi-source approach towards anomaly detection using logs and tracing data. First, we describe the logs and traces as generated by the system. We present their specifics that are exploited for the definition of the Next Template Prediction (NTP) pseudo-task. Second, we describe the NTP pseudo-task for anomaly detection. Thirdly, we describe one way to address the NTP task utilizing deep learning architecture on a single modality description of the system state. Next, we provide a solution that enables us to efficiently solve the NTP problem as a pseudo task for joint detection of anomalies from both logs and traces. Finally, we present an approach that uses the results from the NTP task and performs anomaly detection.

\subsection{Data Representation}
The raw logs and traces as generated by the system, contain various information about the specific operation being executed. Since some of the information is a sporadic description of the operations, proper filtering and representation should be done. Due to the specifics of the two modalities, we address them separately.

\subsubsection{Logs}\label{lofs:preprocessing}
A log is a sequence of temporally ordered unstructured text messages $L=\{l_{i} \,:\,i=1,2,...\}$. Each text message $l_{i}$ is generated by a logging instruction (e.g. printf(), log.info()) within the software source code. Since the logging function is part of the body of the whole program, it can serve as a proxy for the program execution workflow. Hence one can infer the normal execution pattern within the program workflow. 

The logs consist of a constant and a varying part, referred to as log template and log parameters. Due to the large variability of the parameters, they can introduce a lot of noise. To mitigate this problem common way to represent the logs is with the extraction of the constant part through a log parsing procedure. It allows for the creation of a dictionary of log templates from a given set of logs.

To unify the representations of the logs, the log templates are tokenized. A dictionary from the tokens, representing the vocabulary of all of the tokens in the logs - $D_{logs\_words}$ is created. Since the log templates can have a different number of tokens, for the uniform representation of the log templates a special \texttt{<SPECLOG>} token is added, such that each of the logs has an equal number of tokens. The maximal size of the log template is limited by a parameter called $max\_log\_size$.
\begin{equation}
    L_{i}=\{W_0^i, W_1^i, \dots, W_t^i\}
\end{equation}
where each of the $W_t$ is an extracted word mapped to index $t \in D_{logs\_word\_indecies}$.

\subsubsection{Distrubted traces}
\sloppy{Distributed traces are a request-centred way to describe behaviour within the distributed system. It means that they follow the execution of the user issued a request through the distributed system in a record referred to as spans. The spans represent information (e.g. start time, end time, service name, HTTP path) about the operations performed when handling an external request in service. Formally, a trace is written as}

\begin{equation}
    T_{i}=\{S_0^i, S_1^i, \dots, S_m^i\}
\end{equation},
where $i\in \{1,\dots, N\}$ is a trace as part of an observation set of traces, and $m$ is the length $T_i$ or the number of spans in the trace. 

One of the most characteristic properties of the spans is the function executed during the event and a corresponding endpoint. They usually represent either HTTP or RPC calls, denoting the interconnection between the spans within the trace. The HTTP calls are described with \textit{path, scheme, method}. The RPC calls are represented with the functions they are executing. Since these features represent the intra-service communication in a trace, we assume that they are sufficient for structural analysis of possible anomalies. To provide a richer representation of the traces, further augmentation of the traces can be done. More specifically, two artificial spans (\texttt{<START>} and \texttt{<END>}) are added to the beginning and the end of the trace, accordingly. It preserves the knowledge for the length of the trace.

Represented in this form the spans have very similar representation as to the logs, with additional constraints that the spans are further bounded by the operation executed within the trace. It means that they also are facing the problem of the presence of noise into the representation induced by the varying parameters. Similar as for the logs, applying a template extraction technique produces a set of representative template spans. It allows for each of the trace to be represented as a sequence of template spans. Formally,

\begin{equation}
    T_{i}=\{St_0^i, St_1^i, \dots, St_k^i\}
\end{equation}
where each of the $St_k$ is an extracted template mapped to index $k \in D_{template \ indecies}$.

Observing that each function calls are sequences of characters, a dictionary of the sequences of characters appearing inside the given set of traces is constructed - $D_{span \ words}$. It provides a unique language for the description of all of the spans appearing in the observed traces. Formally a span is represented as

\begin{equation}
    St_{j}=\{W_0^i, W_1^j, \dots, W_q^j\}
\end{equation}
where $W_q$ is a sequence of characters as extracted from the dictionary of span words $D_{span \ words}$. Since there are spans with a different number of words, to provide spans in an appropriate representational format for later processing, each of the spans is augmented with a \texttt{<SPECSPAN>} token.


\subsection{NTP: Pseudo-task for Anomaly Detection}
Representation of both traces and logs in the previously described manner, allow us to take a unified approach towards their modelling. The appearance of the next log message is conditioned on the appearance of the history of the previous logs. Similarly, within a trace, the appearance of the next span is conditioned on the previous ones. 
Thus the modelling problem can be conceptualized formally as

\begin{equation}
    P(A_{T_{win}:T})=\prod_{t=T_{win}}^{T}P(A_{t}\vert A_{<t})
\end{equation}
where $A_{<t}$ denotes the templates traces or logs from $A_{t-win}$ to $A_{t}$, with $win$ denoting the size of the preserved history. Hence we refer to this task as the next template prediction (NTP). 

\subsection{Single Modality Anomaly Detection}
\begin{figure}[t]
\centering
\includegraphics[width=1.\textwidth]{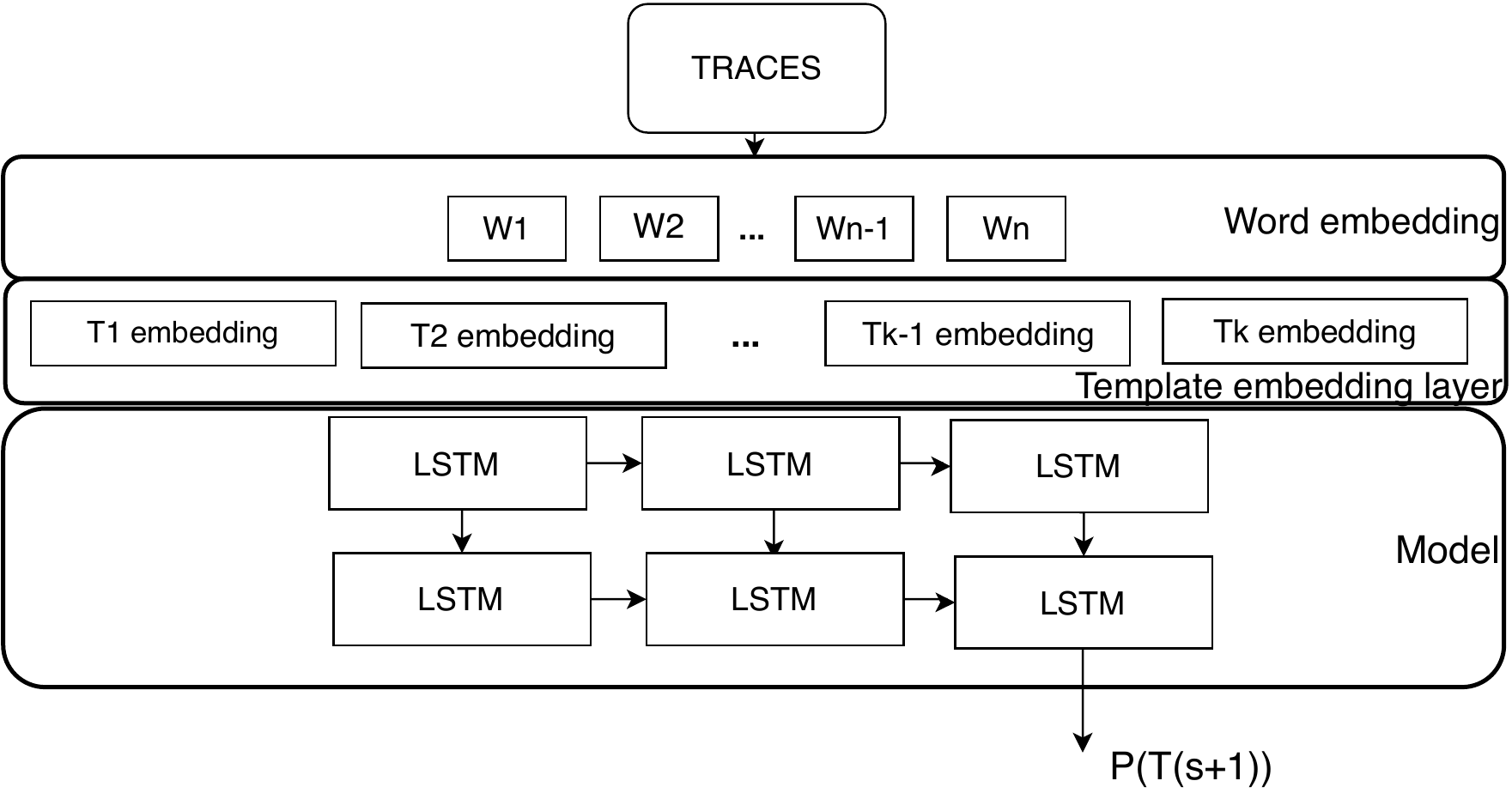}
\caption{Proposed architecture for single modality. The same approach can be utilized also for the logs data.}
\label{fig1}
\end{figure}

\figurename~\ref{fig1} depicts the proposed end to end architecture to solve the NTP task for single modalities. We use the same architecture for both the logs and the traces.

At the input, we provide the dictionary of the words as appearing in $D_{logs\_words}$ and $D_{span\_words}$. We perform initialization with random vectors for each of the words with a specific size. This is a parameter of the method referred embedding size $N_{embedding}$. The template embedding layer uses the representations of the words to create the corresponding sequences of templates. These sequences are fed through an autoregressive deep learning LSTM method that is modelling the sequential dependence between the input samples represented with $f(x)$. Its output is used to calculate the softmax between the real next template and the output of the network. The softmax is calculated as
\begin{equation}
    P(f(x))=\frac{e^{f(x)}}{\sum\limits_{i=1}^A e^{f_{i}(x)}}
\end{equation}
It calculates a distribution over the all possible templates. The one with the maximal probability is considered the most likely template to appear given the input sequence of templates.

LSTM architecture is a deep learning neural network method used for efficiently modelling sequential data. The representation of the system state is given via a single vector, refer to as a hidden state. The assumption the method is making, builds on top of the Markov property. It states that the state of the system at any particular point in time can be determined just from the previous state. To achieve this goal, it utilizes a selection mechanism build on abstractions of input, output and forget gates. This mechanism allows the network to selectively choose how much information from the previous inputs it should preserve and distribute towards the output. Hence it can model short and long term dependencies within a sequence and the structure appearing into the sequence of state events. Thus it is a handy solution for modelling our problem. Stacking of multiple LSTM cells provides greater representational power of the architecture.

\begin{figure*}[tb]
\centerline{\includegraphics[width=\textwidth]{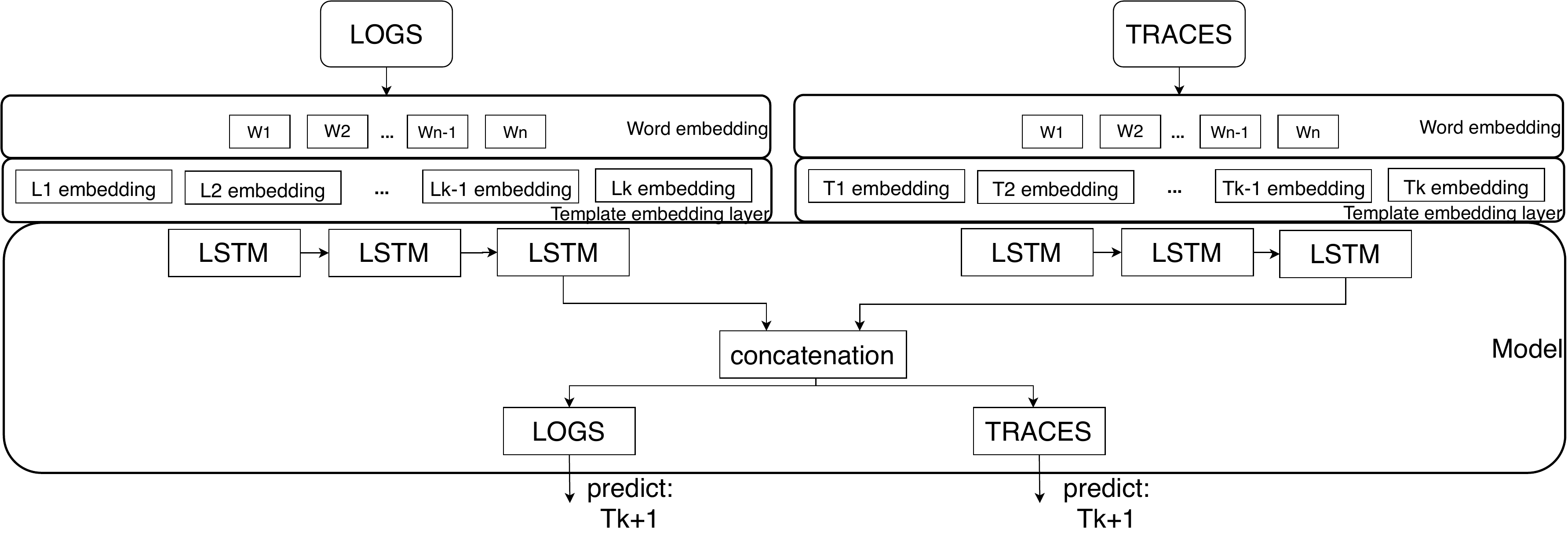}}
\caption{Proposed architecture for joint analysis of logs and traces.}
\label{fig2}
\end{figure*}
\subsection{Multimodal LSTM}

To account for both modalities and enable end to end learning system for anomaly detection, we propose the method as given on \figurename~\ref{fig2}. It is composed of two models  described in the previous section. On the inputs provided are the dictionary of logs and spans, simultaneously, to each of the two models. However, the output of both LSTMs is concatenated to one another and fed through an additional linear layer. It gives an advantage of including the information from both of the modalities, to improve the predictive performance. The shared information from the concatenation is then passed through two linear layers, one accounting for the traces and the other for the logs. 

To account for both modalities the cost function is also changed. We calculate it as a joint cross-entropy loss of the most likely span and log to appear, given the joint information in a particular period. We calculated the joint loss as follows:

\begin{equation}
    L((s, l), f(x, y)) = L(f(x), s) + L(f(y), l)
\end{equation}
where $L(\cdot, \cdot)$ account for the categorical-cross entropy loss, and $s$ and $l$ for the ground truth span and log templates that should appear as the next relevant templates.
Because the loss function includes the information from both modalities when the back-propagation step is done the gradients are calculated based on the information from both of the modalities.

One important detail for joint training the two modalities is providing the information from the same time intervals to the model from both of the modalities. The granularity representation of a log message is on a single time interval, on one side, and the spans span across multiple time stamps. To address this challenge we address block of logs of varying size. The size of a block of log messages is dependent on the corresponding spans within the trace appearing during the particular time interval. To create a block of log messages we stack multiple logs together to pair up with the corresponding time intervals determined by the spans. Such an approach requires the introduction of a maximal number of logs that are considered at once. 

Given this coupling between the traces and logs, the question to ask is "What is the learning task for the joint method?". Since the time spanning of the spans determine the size of log blocks, just a $window\_size$ parameter on the traces imposed is. This parameter determines the number of spans the method should use to produce the next one. The block of log messages is created in a way that, the log messages that come from the start time of the first and the end time of the last span in the window of spans are joined into one block. The target is to predict the next expected log. An additional complication that can arise is the absence of logs in a particular time frame. To address this, we denote those windows that have a missing target and drop them from the learning set.

\subsection{Anomaly Detection}
NTP is utilized for anomaly detection for logs, however, the anomaly detection in the traces require additional anomaly detection procedure. We further provide a simple and effective method that acts on the output from the NTP solver to detect if there is an anomaly or not. The anomaly detection procedure for the single modality log model considers a log as normal if the prediction for the log is in the next $top\_k\_logs$. Otherwise, it is predicted as an anomaly. 

For the detection of anomalous trace, the decision procedure should take into consideration the correct prediction among all of the spans in the trace subject to prediction. A span is correctly predicted if, for a given input sequence of spans, the true span is in the $top\_k\_span$ ranked spans. For each trace, this procedure creates an accumulation of the correctly predicted spans. The ratio of incorrectly predicted spans (span error rate) $\frac{num\_err}{length(trace)}$ is considered as an anomaly score for the trace. Setting a threshold on this score can be used for anomaly detection. Finally, for the joint multimodal method, a combination of the previously described techniques is utilized.

\section{Experiments and results}
In this section, we first describe the experimental design we used for evaluation. Second, we provide a detailed analysis of the results from the experiments to justify the improvements the joint information provides. Finally, we discuss the span2vec embedding as a consequence and further contribution of this work.
\subsection{Experiments}

\subsubsection{Dataset preprocessing details}
In the experiments we used the publicly available dataset \footnote{\url{https://zenodo.org/record/3549604}} covering the trace and logs as monitoring components in overlapping time intervals. To the best of our knowledge, this is the only available dataset suited for multi-modal anomaly detection in distributed systems and as such it is utilized.

The experiments are generated from an OpenStack deployment testbed. We used the concurrent execution scenario, with 3 execution workloads: create an image, create a server, create a network, as described in~\cite{Nedelkovski2020Dataset}. As such we demonstrate the usefulness of our method in scenarios as close to real-world execution.

\subsubsection{Train test split}
The training dataset is composed of the traces appearing up to a particular time point, such that 70 \% of the normal traces are contained. The anomalous traces during this time-window are discarded.  The logs that belong in the corresponding time intervals as generated by the trace are also preserved in the training set. We aim of modelling the normal behaviour of the system with preserving the normal traces and normal logs. To evaluate our model, the test set is composed of all of the remaining logs and traces appearing after the split time point. 

\subsubsection{Baselines}
The main aim of this work is to demonstrate that the shared information between the logs and traces can improve anomaly detection in comparison to anomaly detection methods build from single modalities. As baselines we use the single modality LSTM method build separately for the traces and logs. The models are built on the same dataset as the multi-modal model and tested on the same test set to allow for a fair comparison.

\subsubsection{Implementation details}
The first step of the data preprocessing requires settings the values for the Drain parser. The values for the similarity and depth were set to 0.5, 0.4 and 4, 4, for the logs and traces accordingly. These values provide a concise template as evaluated by the domain expert.
The $N\_embedding$ is set to 256.
For the $window\_size$ parameter for the traces the value is set to 3. 
For optimization of the cost functions for the single and multiple modalities methods, we use SGD solver with standard values for the $learning\_rate = 0.001$ and $momentum = 0.9$. The $batch_size$ is set to 256 as a commonly chosen values. The number of $epochs$ is 100 for all of the tested methods.

For the anomaly detection procedure we further require the $logs\_top\_k$ and $trace\_top\_k$ parameters. They are set to 20 and 1 accordingly. For the error threshold on the anomaly score, the best value between 0.05 and 1 with a step of 0.05 chosen is.

\subsection{Results}

\tablename~\ref{tab:results} summarize the results from the experiments. Firstly, one can observe that the results from the single modalities methods show that for the logs and traces, individually the approach can provide good results. It shows that the assumption made by the NTP task solver is sufficient for successful modelling of the normal state of the system.

\begin{table}[!t]
\renewcommand{\arraystretch}{1.3}
\caption{Results from the experimental evaluation.}
\label{tab:results}
\centering
\begin{tabular}{lcccc}\hline
score  & Logs-joint & Trace-joint & Single logs & Single traces \\ \hline
accuracy             & 0.976      & 0.990       & 0.974       & 0.955 \\ 
precision            & 0.904      & 0.992       & 0.897       & 0.992 \\ 
recall               & 0.996      & 0.984       & 0.996       & 0.909 \\ 
f1                   & 0.948      & 0.988       & 0.944       & 0.949   \\ \hline
\end{tabular}
\end{table}

Comparison of the results from the columns Trace-joint and Trace-single suggest that there is an improvement of the results for the traces for the multimodal method. More specifically, there can be observed improved value on the recall for the joint model for the traces in comparison to the single one. This suggests that the addition of the additional information from the logs can increase the number of correct predictions for the anomalous traces. The improvement is further depicted in the increased value for the F1 score on the joint traces. The results on the logs do not seem that change too much. One explanation of this behaviour is that the granularity of the information from the logs is truncated on the level of the data source with a lower frequency of generation - the trace is harder for the information in the trace to be transferred to the logs. The information that the multimodal method is receiving from the logs when it is aiming to predict the next relevant span complements the information as obtained just from the sequence of spans individually.

\subsection{Span2Vec and Log2Vec}
One element of the method is the ability to learn to embed both the logs and spans. The logs and spans are composed of words represented as vectors. The vectors are learned during the optimization procedure. Hence are optimized for the specific NTP task. Since the logs and spans are linear combinations from these words, pooling over the words belonging to the same span/log can be used to provide a unique vector mapping for them.

\begin{figure}[t]
\centering
\includegraphics[width=0.9\textwidth]{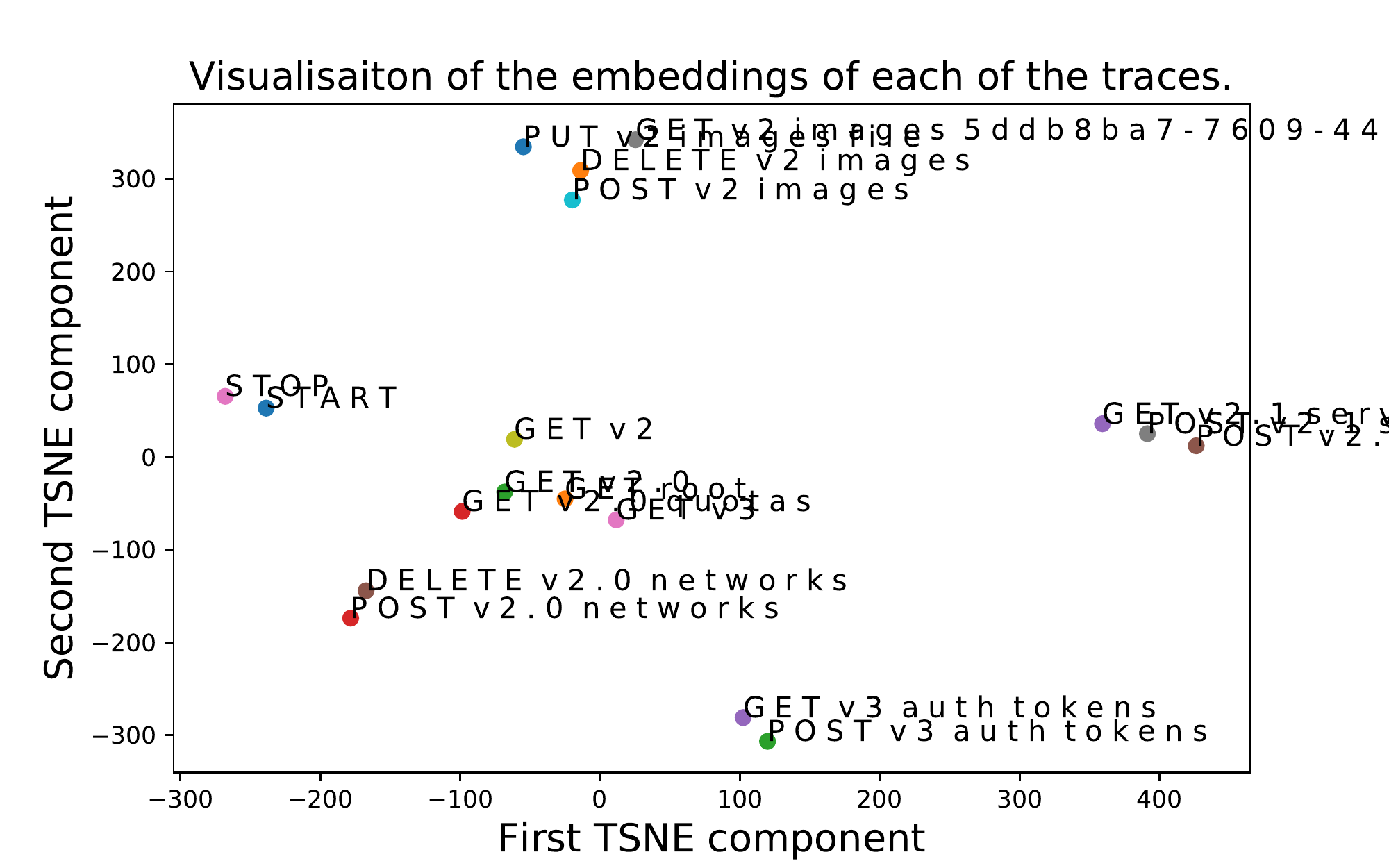}
\caption{Span2Vec embedding of the events in the tracing data from the whole vocabulary of spans for the three different workloads.}
\label{embeddingsspans}
\end{figure}

\figurename~\ref{embeddingsspans} depicts a two-dimensional representation of the vector space of the spans embeddings. Three operations are executed. Close observation reviles that spans that are specific for a workload occur close to one another, while the ones that are shared co-occur in groups of their owns. For example, the spans \textit{GET /v2.0/images/},  \textit{PUT /v2.0/image/}, \textit{GET /v2/images/} and \textit{POST /v2.0/networks/} are unique for \textit{create delete image} workload. As it can be observed, these spans are very close to one another in comparison to the other spans like the pair \textit{POST /v2.0/networks} and \textit{DELETE /v2.0/network/}. On the other side, the artificially added spans like \textit{START} and \textit{STOP} or the authentication span each of the workloads is utilizing are grouped, separated from the workload-specific spans.
Close inspection of the Euclidean distance between the spans confirms the observations from the TSNE vector representation. 
The importance of these embeddings is the most emphasised in their future reuse for warm starting the methods. This can reduce the adoption time and the difficulty when a new machine model is deployed in production.

\section{Conclusion}
In this work, we presented a novel method for multi-source anomaly detection in distributed systems. It uses data from two complementary different modalities describing the behaviour of the distributed system - logs and traces. We utilize the next template prediction (NTP) task as a pseudo task for anomaly detection. It is based on the assumption that the relevant information from the program execution workflow can be preserved into one vector. Then it uses the corresponding vector to predict the most relevant template to appear.
To detect the anomaly, a post-processing step that acts on the predictions of the NTP task is used.

The results show that the multimodal approach can improve the scores for anomaly detection for multiple modalities in comparison to the single modalities of logs and traces. The information that the logs and traces are preserving is complementary and the model can exploit it. Furthermore, the method can produce vector representation for both the logs and traces. These vector embeddings are used as a good bias for transferring and reusing the accumulated knowledge for faster training and adaptation.

In future work, we would investigate how adding additional information from the metric data can be incorporated into the model. It will allow for the creation of a unified model of the whole system behaviour, making the further processes of AIOps life-cycle easier. Additionally, we would investigate transfer learning approaches based on the generated embeddings. Specifically, we are interested in investigating how the learned embeddings can be reused for other types of workloads with a final aim to reduce the deploy time of the machine learning model in production. 

\bibliographystyle{splncs04}
\bibliography{main}
\end{document}